# AGSPNet: A framework for parcel-scale crop fine-grained semantic change detection from UAV high-resolution imagery with agricultural geographic scene constraints

Shaochun Li[a, b, c], Yanjun Wang[a, b, c, *], Hengfan Cai[a, b, c], Lina Deng[a, b, c] and Yunhao Lin[a, b, c]

[a]Hunan Provincial Key Laboratory of Geo-Information Engineering in Surveying, Mapping and Remote Sensing, Hunan University of Science and Technology, Xiangtan 411201, China

[b]National-local Joint Engineering Laboratory of Geo-spatial Information Technology, Hunan University of Science and Technology, Xiangtan 411201, China

[c]School of Earth Sciences and Spatial Information Engineering, Hunan University of Science and Technology, Xiangtan 411201, China

[*]Corresponding authors. E-mail addresses: wangyanjun@hnust.edu.cn (Y.Wang)

**Abstract:** Real-time and accurate information on fine-grained changes in crop cultivation is of great significance for crop growth monitoring, yield prediction and agricultural structure adjustment. Aiming at the problems of serious spectral confusion in visible high-resolution unmanned aerial vehicle (UAV) images of different phases, interference of large complex background and "salt-and-pepper" noise by existing semantic change detection (SCD) algorithms, in order to effectively extract deep image features of crops and meet the demand of agricultural practical engineering applications, this paper designs and proposes an agricultural geographic scene and parcel-scale constrained SCD framework for crops (AGSPNet). AGSPNet framework contains three parts: agricultural geographic scene (AGS) division module, parcel edge extraction module and crop SCD module: (1) AGS division module uses multi-source open geographic data products to delineate AGS with relatively consistent geographic element conditions by analyzing the rule of agricultural territorial differentiation. (2) The parcel edge extraction module uses a bi-directional cascade network (BDCN) and a designed edge optimization model to obtain comprehensive AGS farm parcels. (3) The SCD module uses a designed criss-cross-attention network (CCNet) with pseudo-Siamese structure and change feature discrimination module for extracting semantic features and change features of AGS crops, outputting accurate pixel-level semantic change maps, and then fusing the parcel extraction results with the semantic change maps to finally obtain parcel-scale fine-grained SCD results of crops. Meanwhile, we produce and introduce an UAV image SCD dataset (CSCD) dedicated to agricultural monitoring, encompassing multiple semantic variation types of crops in complex geographical scene. We conduct comparative experiments and accuracy evaluations in two test areas of this dataset, and the results show that the crop SCD results of AGSPNet consistently outperform other deep learning SCD models in terms of quantity and quality, with the evaluation metrics *F1-score*, *kappa*, *OA*, and *mIoU* obtaining improvements of 0.038, 0.021, 0.011 and 0.062, respectively, on average over the sub-optimal method. The method proposed in this paper can clearly detect the fine-grained change information of crop types in complex scenes, which can provide scientific and technical support for smart agriculture monitoring and management, food policy formulation and food security assurance.

change detection; deep learning

**1. Introduction**

Crop spatial distribution mapping is an important information base for the study of land cover changes, regional crop growth monitoring, yield estimation and planting structure adjustment (Wu et al. 2023), which is of great significance for the national formulation of agricultural economic development planning and improvement of agricultural production management (Cai et al. 2018). To rapidly and accurately obtain information on the structure of crop cultivation and ensure the security of food production and supply, it is necessary to detect the dynamic distribution changes of crops (Woodcock et al. 2020). Traditional agricultural investigation through field collection of crop planting information for statistics and aggregation, this method in a large area of the survey work requires extremely expensive labor and material resources, and its results have a lag, cannot be real-time monitoring of agricultural conditions. Remote sensing technology, with its wide coverage and short detection period, has become an important measure for accurate crop change detection (Gerhards et al. 2019). Crop remote sensing change detection (CD) aims to identify fine-grained change information of crop cultivation types by quantitative or qualitative analysis of remote sensing images (RSIs) of the same geographical area acquired at different phases (Liu et al. 2022).

Traditional pixel-based CD methods mainly use image algebraic calculations or spatial transformations to obtain change information of different temporal images (Hussain et al. 2013). Although this method is simple and fast, it is mostly focused on lower resolution multispectral images and cannot be adapted to high-resolution images with greater image element variability, which can produce a large number of pseudo-variable regions and "salt-and-pepper noise". To overcome the above difficulties, machine learning (ML) manual features are applied to CD tasks. For example, Zerrouki et al. (Zerrouki et al. 2019) proposed a weighted random forest algorithm to identify land cover changes in high-resolution satellite images. However, ML methods require a large set of specific manual samples and statistical features, and their shallow model structures are difficult to extract deep features, making them less generalizable to other non-specific regions (Peng et al. 2020). Additionally, satellite RSIs also have many disadvantages, such as longer revisit cycles, lower spatial resolution, and harsh weather conditions such as clouds and fog in agricultural regions (Liu et al. 2021), which can adversely affect the data quality and make it difficult to meet the practical application requirements of large-scale area fine-grained crop dynamics monitoring. Compared with satellite remote sensing, unmanned aerial vehicle (UAV), which has hardware advantages such as high flexibility and low cost, can acquire very high-resolution RSIs in real time and is not constrained by geographical environment, and has been widely applied to crop yield estimation, growth and disease monitoring, etc (Mesquita et al. 2019). However, traditional CD methods focus only on the region of no change and change between different temporal RSIs, namely binary change detection (BCD), which greatly limits its application to multi-category change detection in crops. In contrast to the BCD task, semantic change detection (SCD) not only provides fine-grained change category information, but also obtains "from-to" semantic change information from the change map that includes the direction of change, such as "from rapeseed to rice". This more comprehensive and detailed category change information is essential for fine-grained monitoring of complex crops (Kalinicheva et al. 2020).

With the rapid development of artificial intelligence technology and high performance computers, end-to-end deep learning (DL) networks, represented by convolutional neural networks (CNN), are flourishing (Shafique et al. 2022). DL methods have powerful deep feature extraction capabilities, can easily generate change maps of different temporal images, and have become the mainstream method for high-resolution image CD in recent years (Shi et al. 2021). The traditional deep learning SCD methods mainly adopt the direct classification (DC) method (Peng et al. 2019) and the post-classification comparison (PCC) method (Wu et al. 2017). The DC method first fuses the corresponding pixels of different bands of bi-temporal images into a single multi-band image by superimposing or differencing them, then assigns a unique semantic change information to each pixel of the image and serves as the input of semantic segmentation networks (such as U-Net and DeeplabV3+, etc.), finally outputting the semantic change map. This method tends to ignore the diversity of surface cover change features on high-resolution images with rich feature details. The PCC method classifies the bi-temporal images separately using a semantic segmentation network, and then distinguishes the final change regions and directions by comparing the classification maps of the bi-temporal images. This method is simple and intuitive, but the instability of the classification map accuracy may cause the accumulation of errors in the final SCD results. Furthermore, it is unable to exploit the temporal dependence that is significant for the CD task to obtain discriminative bi-temporal change features. To solve the above problems, Daudt et al. (Daudt et al. 2019) constructed a multi-task learning strategy by integrating a Siamese full convolutional network (FCN) for extracting semantic features and a separate FCN for extracting change features. They combined two weighted loss functions to enable the Siamese FCN and the separate FCN to perform semantic segmentation and BCD, respectively, to generate a categorical map containing change types and a binary change map containing change regions, respectively, finally assigning change directions to the change regions and obtaining the final semantic change map through post-processing.

Nevertheless, the current application of SCD methods to high-resolution RSIs for agricultural monitoring are still haunted by several problems: (1) confusing spectral visual features (Zhu et al. 2022a). Due to the difference of imaging conditions in different phases, the changed regions may show strong spectral similarity, for example, it is difficult to distinguish between rapeseed and rice grown in different phases whether they are changed or not, while the unchanged regions may also produce large spectral differences, causing the SCD methods to incorrectly detect the unchanged regions as changed regions. (2) "Salt-and-pepper" noise (Yu et al. 2016). The current state-of-the-art deep learning SCD methods are still mainly based on semantic segmentation algorithms, which tend to classify each pixel of the farmland independently, and thus may generate a large amount of "salt-and-pepper" noise, making it difficult to render the complete parcel boundaries. (3) Complex background features with unbalanced proportions (Zhu et al. 2022b). Change areas often occur only in a portion of the region, while the unchanged background with a disproportionate proportion often shows complex distribution characteristics of non-agricultural areas and may have similar characteristics with the change area ground features, which can make the SCD method strongly interfered in the detection. To reduce these problems, some scholars have considered object-based image segmentation methods to aggregate homogeneous features such as spectra, textures and spatial contexts to form image objects bottom-up, such as multi-scale segmentation method (Liang et al. 2022). However, these methods are essentially only shallow feature aggregation and do not fully utilize the deeper features of the image, and their segmented object units often do not match the actual target feature morphology that one needs.

Thanks to the DL network, it can simulate human vision to extract the high-level features of the images and distinctly depict the complete farmland parcel boundaries. For example, Liu et al. (Liu et al. 2020) employed the Richer Convolutional Features (RCF) edge detection network to extract mountainous farmland parcels from high-resolution optical images and time-series synthetic aperture radar (SAR) images, and gained a high accuracy farmland parcel distribution map. Although the DL edge extraction algorithm has been considered as the most promising method for depicting agricultural parcels (Sun et al. 2022), it still suffers from complex background features over large areas in practical agricultural applications. The reason may be that the DL algorithm is only mechanically improving the internal structure of the network (such as attention mechanism (Li et al. 2020), inflated convolution (Fan et al. 2022) and loss function optimization (Wang et al. 2022), etc.) without considering the complex geographic environment and high-intensity spatial heterogeneity existing in the images, and it is hard to effectively construct a complete mapping relationship of crops from image space to geographic space using a single DL model, and thus cannot match the rule of territorial differentiation (Myint et al. 2011). In recent years, geographic scene division has increasingly become an effective solution to the problem of large and complex scenes (Xu et al. 2019). For example, Sun et al. (Sun et al. 2020) proposed a hierarchical perception approach in which they used a road network to divide the images into multiple geographic blocks and applied scene constraints to the extracted geographic entities, and then performed crop classification by a DL model to obtain better classification results while reducing the scene complexity. The above studies show that adopting effective constraints on geographic scenes and parcel boundaries can contribute to improving the dynamic distribution results of complex multiple crops on deep learning SCD applications.

To sum up, we propose an agricultural geographic scene constrained parcel-scale semantic change detection framework (AGSPNet) for detecting fine-grained semantic change information of crops in bi-temporal UAV RSIs. The proposed AGSPNet mainly comprises three modules, namely, the agricultural geographic scene division, parcel edge extraction, and crop SCD network, to solve the problems of difficult crop change feature extraction for bi-temporal high-resolution RSIs and mismatch between farmland detection edges and reality. Therefore, the main contributions of this study are summarized as follows.

(1) We design an agricultural geographic scene division module for the UAV RSIs crop SCD task in complex scenarios. This module adopts multi-source open geographic data to geographically analyze UAV imagery and distinguish agricultural geographic scene with relatively consistent geographic element conditions and terrain features, thus mitigating the influence of complex background features of large non-crop areas.

(2) We extract large-scale complete farm parcels based on a bi-directional cascade network (BDCN) and a designed parcel edge optimization model for constraining crop semantic change results, thus making the final crop semantic change map to match the real demand.

(3) We provide a new SCD module for the bi-temporal UAV RSIs crop SCD task. This module constructs the cross-cross-attention network (CCNet) as a pseudo-Siamese structure and incorporates the change feature discrimination module, which improves the model's ability to extract bi-temporal crop change features and maximizes the potential performance of SCD networks for crop mapping.

(4) We create a new SCD open access dataset, namely CSCD, dedicated to large scale yearly agricultural monitoring. This dataset is based on UAV high-resolution RSIs with clearer bi-temporal multiple crop change types and "from-to" semantic change annotation information, which can provide effective data support to advance the research of new SCD methods. The CSCD dataset is available at https://doi.org/10.6084/m9.figshare.22561537.v1.

## 2. Materials and methods

*2.1. Study area*

The study area locates in Yuhu District, Xiangtan City, Hunan Province, China, as shown in Fig. 1 (a), and mainly contains two experimental areas with a total area of 20.65 km², of which the training dataset area is 16 km² and the testing dataset area is 4.65 km². Yuhu District locates in the eastern part of Hunan Province, and its northern region is full of hills and mountains with a dense network of water. It is a subtropical monsoonal humid climate with abundant heat and rainfall, annual average temperature between 15~23 ℃ and annual precipitation of 1320 mm, its good water and heat environment is suitable for growing many kinds of cash crops and food crops, such as rapeseed and rice in many seasons. However, the interlocking topographic distribution and complex cropping structure of the region also pose great challenges for SCD of crops.

*2.2. Crop semantic change detection dataset*

*2.2.1. Dataset description*

To validate the performance of the proposed framework for agricultural monitoring, we produced a new yearly bi-temporal UAV high-resolution image crop SCD dataset (CSCD). The images were collected on April 15 (T1 phase) and October 2 (T2 phase), 2021, with a 0.2 m spatial resolution, a spectrum in the visible band (RGB), and an image size of 21952 × 23520 pixels. To facilitate DL network training, we segmented the bi-temporal images into 10290 pairs of 224 × 224 pixel non-overlapping image blocks. The CSCD dataset differs from the existing publicly available SCD dataset in the following aspects: (1) Different spatial resolution. Most of the existing SCD datasets have low spatial resolution, such as Landsat-SCD (Yuan et al. 2022), which cannot meet the data quality requirements of new SCD methods. The CSCD dataset acquires 0.2 m high spatial resolution RSIs based on UAV, which can make the features present very detailed structural information such as texture, geometry and shape, providing the possibility of fine-grained SCD of crops. (2) Different data annotation. Most of the existing SCD datasets only have labels for the types of ground change in different phases (Yang et al. 2021), while the CSCD dataset not only contains bi-temporal crop multi-class change and no-change labels, but also indicates explicit "from-to" semantic change information, providing different model validation spaces for both BCD and SCD tasks. (3) Different applications. The existing scene-level SCD datasets are rich in change types, but their definitions are too broad and often difficult to apply in practical engineering (Peng et al. 2021). The CSCD dataset focuses on agricultural investigation and contains six complex crop change types, which ensures the diversity of data types while expanding the application scope and depth of the SCD dataset. (4) Different time span. The existing SCD datasets have a large time span and are based on a one-year or multi-year time scale (Song & Choi 2020), which cannot provide a available data source for fine-grained monitoring of crops with a short growth cycle. The CSCD dataset shrinks the time scale to within one year and contains bi-temporal RSIs for April and October, which provides support for fine-grained monitoring of crop growth and rotation. Therefore, the proposed CSCD dataset well complements the existing SCD datasets in terms of spatial resolution, time span, change type and application.

*2.2.2. Data preprocessing*

The composite wing UAV (CW-10, Chengdu JOUAV Automation Tech Co., Ltd, Sichuan, China)

that acquired the original images of the CSCD dataset has a wingspan of 2.6 m, a payload weight of 2 kg, an endurance of 90 minutes, a cruising speed of 20 m/s, and a CA-102 full-frame ortho camera. The CA-102 full-frame ortho camera has an Exmor R CMOS sensor type, a sensor size of 35.9 × 24.0 mm, a resolution of 7952 × 5304, contains approximately 42.4 million effective pixels, a memory capacity of 128G, and a lens focal length of 35 mm. The flight height of the UAV was set to 200 m, with a heading overlap of 60% and a collateral overlap of 40%, thus acquiring images in the visible band with 0.2 m spatial resolution. The images preprocessing were done in Pix4D software. First we stitched the acquired images of different UAV sorties, then added the ground control points collected on site for geometric correction, and generated the UAV orthophotos after setting the coordinate system to CGCS2000.

According to the actual survey needs of the local agricultural sector in Yuhu District, the crop types of the CSCD dataset are classified into vegetable, nursery, rapeseed, early-season rice, middle-season rice, and late-season rice in this study. The samples of six typical crop types are shown in Fig. 2, we show rapeseed at maturity and early-season rice at sowing in April, and middle-season rice at harvest and late-season rice at maturity in October. The actual agricultural survey requires counting not only the acreage of crops that change during the year, but also the acreage of important unchanged crops, such as vegetable, which has a short growth cycle and is planted more frequently, and nursery, which grows more slowly. Therefore, we divide the semantic change categories of the CSCD dataset into seven categories, each with a unique "from-to" semantic change label, based on the phenological periods of crops grown in the same area of different temporal images (Fig. 1 (d)). The percentage of area and the number of parcels in the training and testing datasets for different semantic change categories are shown in Table 1.

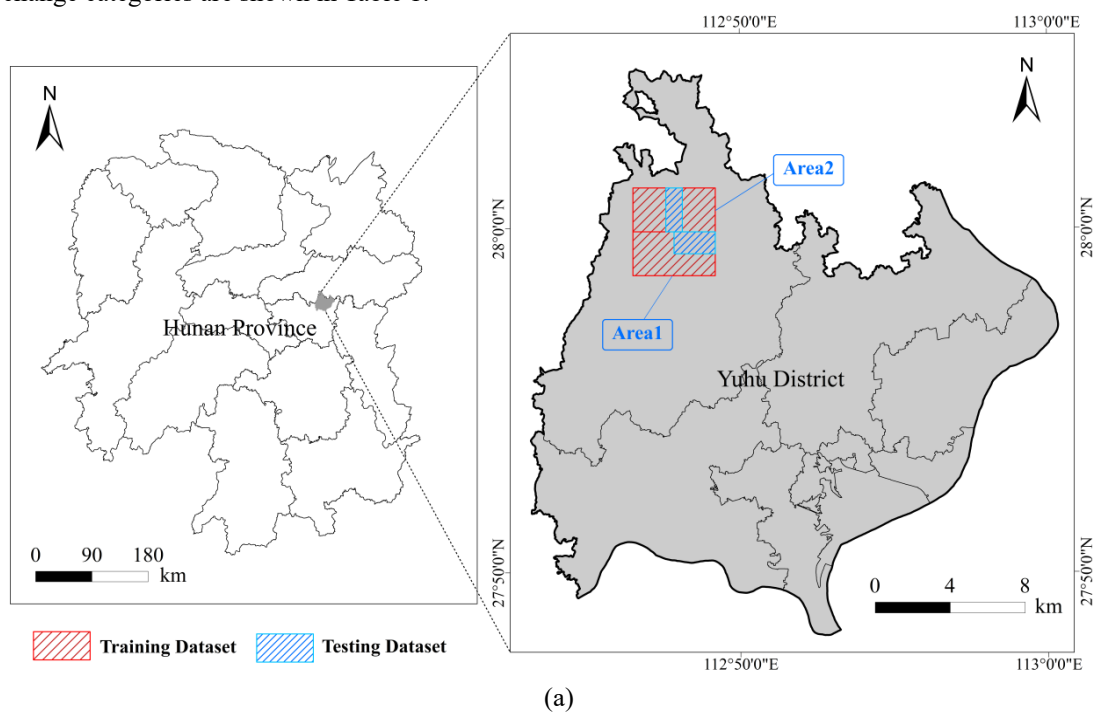

(a)

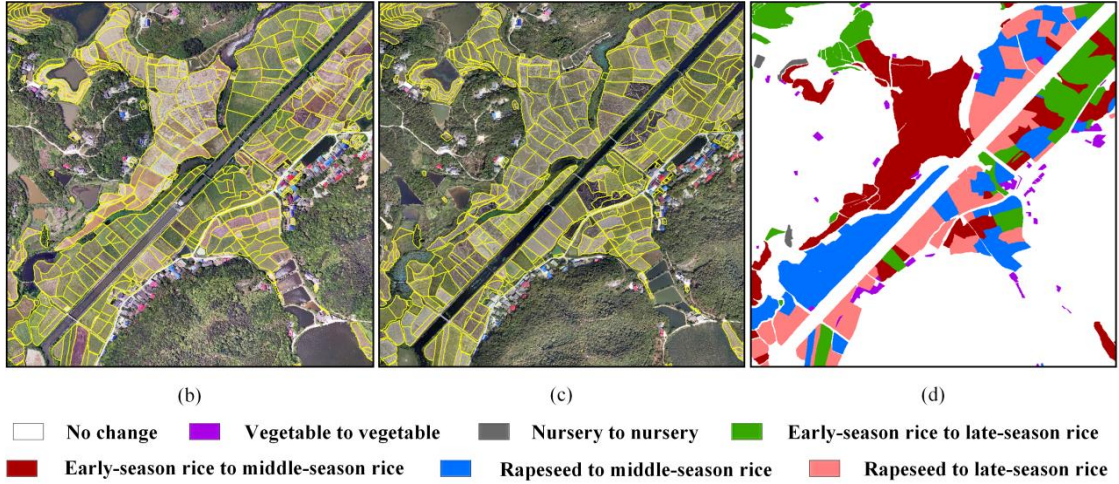

| | | | |
|---|---|---|---|
| (b) | (c) | (d) | |

Legend: No change | Vegetable to vegetable | Nursery to nursery | Early-season rice to late-season rice | Early-season rice to middle-season rice | Rapeseed to middle-season rice | Rapeseed to late-season rice

**Fig. 1.** Study area. (a) Yuhu District locates in Xiangtan City, Hunan Province, China; (b) T1 temporal image sample; (c) T2 temporal image sample; (d) semantic change category sample.

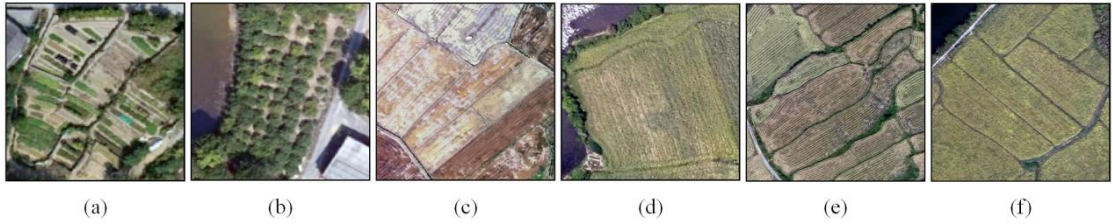

(a)      (b)      (c)      (d)      (e)      (f)

**Fig. 2.** Typical crop types. (a) Vegetable; (b) nursery; (c) early-season rice; (d) rapeseed; (e) middle-season rice; (f) late-season rice.

**Table 1.** Sample profile of semantic change categories (percentage of area (%) / number of parcels).

| Change Type | Training Dataset | Testing Dataset | Total |
|---|---|---|---|
| no change | 59.4/1 | 14.5/1 | 73.9/2 |
| vegetable to vegetable | 1.2/1335 | 0.4/431 | 1.6/1766 |
| nursery to nursery | 0.7/187 | 0.2/78 | 0.9/265 |
| early-season rice to middle-season rice | 4.2/998 | 1.8/362 | 6/1360 |
| early-season rice to late-season rice | 7.8/1613 | 3.6/804 | 11.4/2417 |
| rapeseed to middle-season rice | 1.9/414 | 1.2/226 | 3.1/640 |
| rapeseed to late-season rice | 2.2/439 | 0.9/169 | 3.1/608 |

## 2.3. Methods

### 2.3.1. AGSPNet framework overview

In this study, we propose a framework for SCD of complex crops from UAV high-resolution images with agricultural geographic scene and parcel boundary constraints, namely AGSPNet, and the overall framework is shown in Fig. 3. AGSPNet is mainly composed of three parts: the first part is the agricultural geographic scene division module, which is used for analyzing the rule of agricultural territorial differentiation in bi-temporal UAV images and dividing the bi-temporal agricultural geographic scene images as the input dataset for other modules. The second part is the DL parcel edge extraction module, which is used to further extract and optimize the edges of agricultural parcels from agricultural geographic scene images and serve as boundary constraints for the initial crop SCD results. The third part is the crop SCD module, which is used for the final crop SCD result prediction and post-processing. We will describe the principles of the three modules of the AGSPNet framework in detail in the next few subsections.

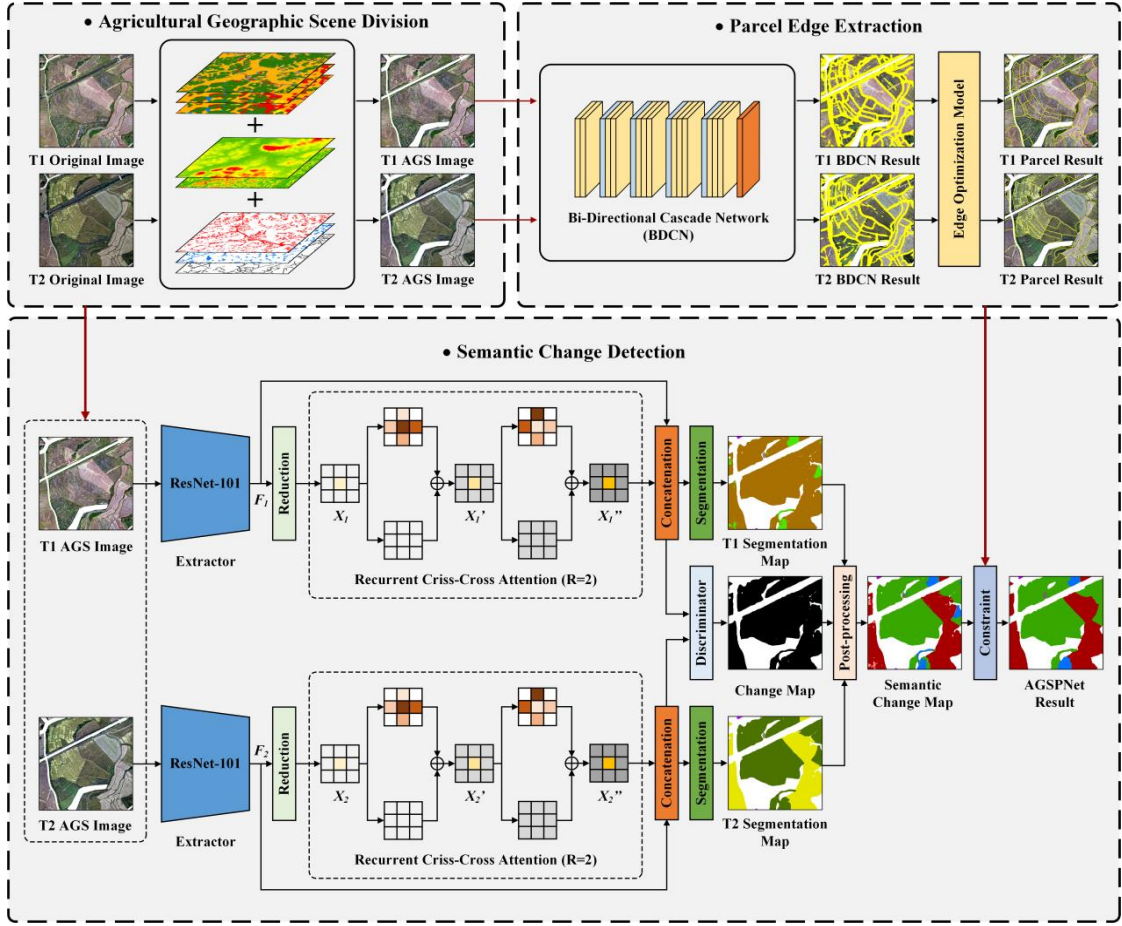

**Fig. 3.** The overall framework of AGSPNet.

*2.3.2. Agricultural geographic scene division*

Rugged topographic features and special hydrothermal environments can produce high-intensity spatial heterogeneity, making crop classification and change detection extremely difficult. As Table 1 shows, the samples of no-change type in the CSCD dataset containing large hilly and mountainous areas occupy 73.9% of the whole study area, and too much background of no-change area will make the DL model more biased to the calculation of background during training. Meanwhile, it often has various complex ground features unrelated to crops, such as mountain, buildings and water body. All the above situations can seriously interfere with the feature extraction of crops by the model. Thus, it is necessary to address the problem of complex backgrounds in large unchanged areas using effective strategies.

Inspired by the first law of geography (Tobler 1970) and the rule of territorial differentiation, that is, the existence of relative consistency of regional characteristics for the same kind of features and spatial heterogeneity for different kinds of features, we design a method for dividing agricultural geographic scene applicable to complex topographic conditions, as shown in Fig. 5. The main idea is to employ multi-source open geographic data products to perform geographic analysis of the study area, and then delineate agricultural geographic scene with relatively consistent topographic conditions and ground features. The multi-source open geographic data products used include global 10 m resolution land use / land cover (LULC) data, ALOS satellite 12.5 m resolution high-precision digital elevation model (DEM) data (https://search.asf.alaska.edu/#/) and Open Street Map (OSM) historical vector data

(https://download.geofabrik.de/asia/chin a.html#). The global LULC data products include the 2017 and 2021 Environmental Systems Research Institute (ESRI) LULC products (https://www.arcgis.com/apps/instant/ media/index.html?appid=fc92d38533d440078f17678ebc20e8e2) and the 2020 European Space Agency (ESA) WorldCover products (https://zenodo.org/record/5571936#). Both data products are derived from global 10 m resolution land cover classification results from Sentinel-1 and Sentinel-2 satellite imagery, with an accuracy of 85% for the ESRI product and 74% for the ESA product, providing a high base accuracy for the delineation of agricultural geographic scene. The DEM data contains not only itself but also slope data after slope analysis. The OSM data includes line and polygon crowdsourced historical vector data of buildings, roads and water bodies.

The main operation flow of agricultural geographic scene division is shown in Fig. 4. At first, considering the regional extent of the bi-temporal UAV images and the local crop cultivation structure, we use ESRI and ESA global land cover products under the same extent. To ensure the complete distribution of crops to the greatest extent possible, we overlay and extract the land cover related to agriculture from the two data products, including trees, crops and scrubs, as a way to obtain pre-selected areas for agricultural geographic scene. Then, we overlay the pre-selected areas with DEM and slope maps for analysis, and obtain the maximum elevation of 93 m and the maximum slope of 16° for crop distribution, and treat the trees and shrubs above these values as non-agricultural areas for elimination. Finally, the remaining non-agricultural areas in the pre-selected areas are further removed using the building, road and water vector data from the OSM data, and the preserved areas are the agricultural geographic scenes. Based on the vector contours of the agricultural geographic scenes, we crop the bi-temporal UAV images and sample data as input datasets for subsequent processing in other modules.

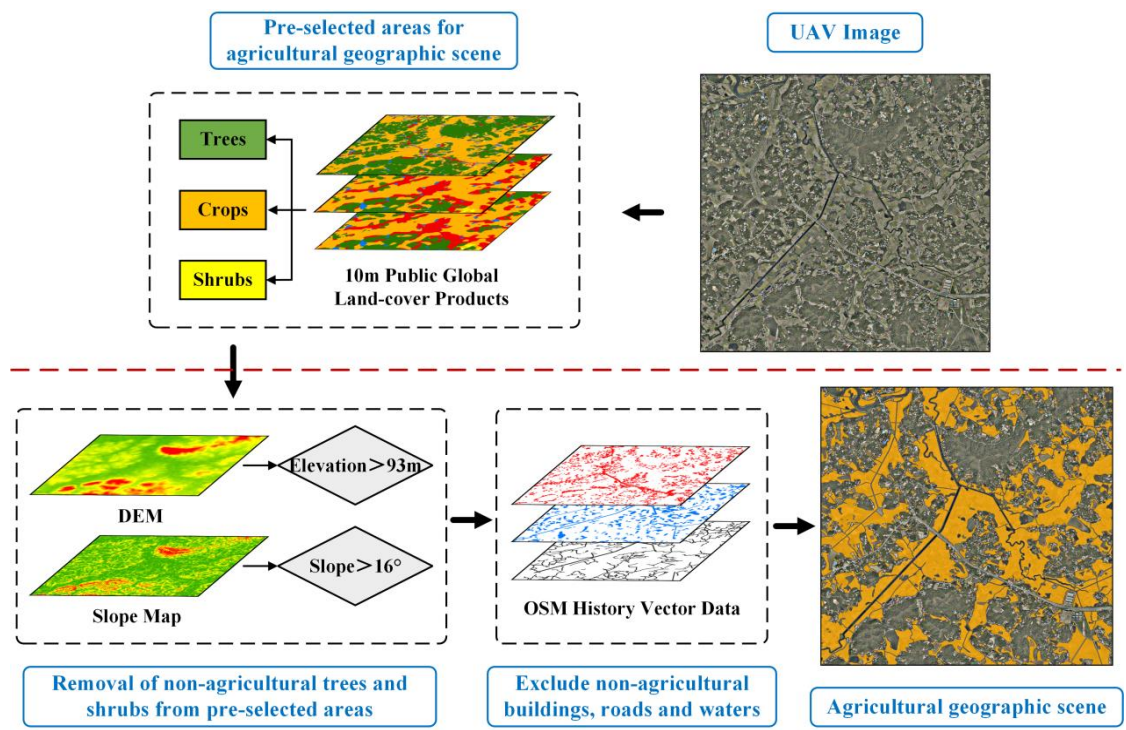

**Fig. 4.** The flow chart of agricultural geographic scene division.

*2.3.3. Parcel edge extraction*

There is often a significant semantic-gap (Zhong et al. 2015) between the pixel-level SCD output of existing semantic segmentation methods and real geographic objects. Although object-oriented image segmentation methods can alleviate the above situation, they also suffer from the drawback of not being able to exploit and express the high-level semantic features of the images to generate objects that match the geographic entities interpreted by manual visual interpretation. Besides, the complex composition around the farmland parcels, such as paths between farmlands, bare soil and vegetation distributed with crops, etc., also poses great difficulties for boundary detection. Thanks to the development of DL edge extraction technology, it can achieve accurate detection of all edge pixels of RSIs by training parcel samples with semantic annotations, and then convert the raster results into polygons through post-processing to obtain complete farmland parcels that conform to the visual perceptual morphology.

This study adopts a bidirectional cascade network (BDCN) (He et al. 2019) to achieve the extraction of farmland parcel edges, which can effectively solve the problem of difficult multi-scale feature extraction in traditional edge detection networks. BDCN is a lightweight VGG-16-based (Patra et al. 2022) network, which consists of multiple incremental detection (ID) blocks. Each ID block is composed of divided VGG-16 convolutional layers inserted into scale enhancement module (SEM), and different ID blocks use different scales of edge markers for supervised learning. SEM uses inflated convolution to enhance the multi-scale feature details in the output of each layer, and finally fuses the outputs of all layers. Due to the complex network structure of BDCN, we show three of the ID blocks and SEM, as shown in Fig. 5, where $r_0$ denotes the expansion rate factor, $K$ denotes the number of inflated convolution layers, $P$ denotes the edge prediction map, $s2d$ denotes propagation from shallow to deep layers, and $d2s$ denotes propagation from deep to shallow layers, and the output fusion process in the direction of $s2d$ and $d2s$ for each ID block is the bidirectional cascade structure. Specifically, first, the images are input to VGG-16 with three fully connected layers and the last pooling layer removed for feature extraction, and the remaining 13 convolutional layers of VGG-16 are divided into five ID blocks, each followed by access to a 2 × 2 pooling layer to gradually expand the perceptual field of the next ID block. Then, the feature maps extracted from each ID block are input to several SEMs and the outputs are fused into two 1 × 1 convolutional layers to generate two edge predictions $p^{d2s}$ and $p^{s2d}$, and to calculate the class-balanced cross-entropy loss function for the current ID block. Finally, the edge predictions of all ID blocks are fused using a 1 × 1 convolution to obtain the final edge prediction results.

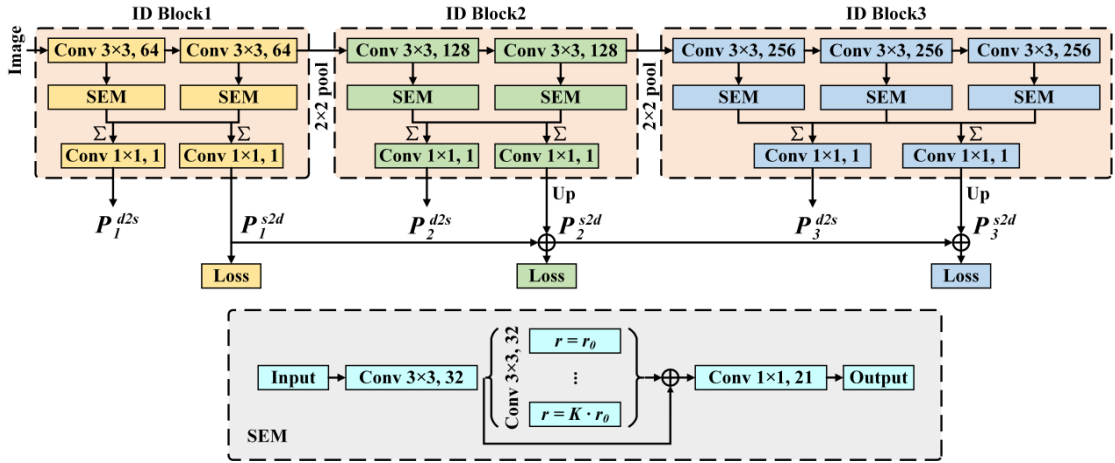

**Fig. 5.** Partial structure of BDCN.

Since the initial output of BDCN is a relatively coarse pixel-level farmland parcel edge map, we require post-processing to obtain a complete farmland parcel at the object level. In this study, we design an edge optimization model to automate batch processing of farm parcel edges using the model builder of ArcGIS 10.6, as shown in Fig. 6. The detailed operation process is as follows: (1) we reclassify the initial BDCN result raster map into a binary map and perform the expansion operation on the edge values; (2) converting raster results to polygons and removing background values and small fragmented parcels; (3) simplifying operations on polygon edges to eliminate voids inside polygons and polygons with empty geometry; (4) converting a polygon to a center line and performing a simplification operation, then extending the line and performing a trimming operation; (5) in the end, the trimmed line elements are converted into polygons, which are the complete farmland parcels at the object level. We fuse the farm parcel results from bi-temporal UAV images, which are used to constrain the boundaries of the initial crop SCD results.

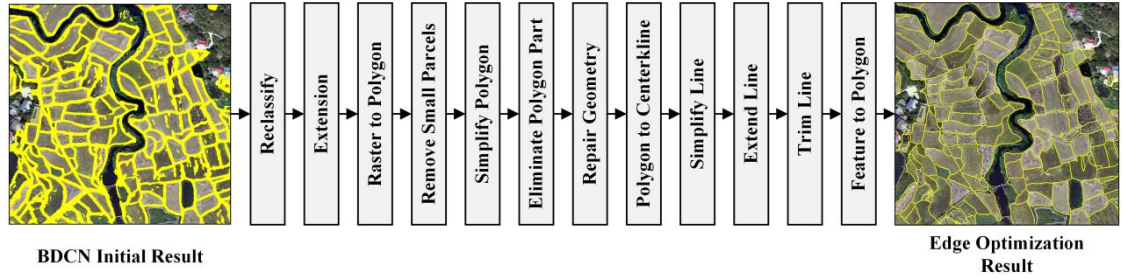

**Fig. 6.** The flow chart of edge optimization model.

*2.3.4. Semantic change detection*

Due to the differences in imaging conditions (Wang et al. 2021a) (e.g., illumination, viewing angle, etc.) in different phases of the UAV, there may be strong spectral similarities or differences between crops in different phenological periods, which is particularly serious in high-resolution visible imagery. Spectral similarity can make it difficult to distinguish crops that have changed in different temporal images, for instance, rapeseed and late-season rice have different semantic features, but they present similar spectral features on the images, which are likely to be identified as unchanged. And spectral difference can likewise cause crops without semantic change to be mistakenly detected as having changed, e.g., vegetables at sowing stage and vegetables at maturity stage. To avoid the above problems, we design a criss-cross-attention network (CCNet) (Huang et al. 2019) with pseudo-Siamese structure to implement SCD for bi-temporal image crops. The network mainly consists of two CCNets that do not share weights and a discriminator FCN, the former is used to extract semantic features of crops in different temporal images and the latter is used to extract bi-temporal change features. The pseudo-Siamese structure (Li et al. 2022b) allows different branches of CCNet to perform better in crop semantic segmentation without the limitation of shared parameters, and also provides help in detecting change regions. The detailed structure of the designed SCD framework is described as follows.

**Feature Extractor.** The feature extractor of CCNet uses a pre-trained ResNet-101 network (Wang et al. 2021b), which can solve the deep network degradation problem that cannot be handled by traditional DL networks, and extract complex deep semantic features more effectively. In order to output feature maps with the same size as the input image, we remove the global pooling layer and the fully connected layer of the network. The ResNet-101 network extracts shallow image features and resizes the feature map to half of its original size by using a $7 \times 7$ convolutional kernel with a step size of 2 and a maximum pooling layer, and then employs four residual blocks to extract higher-level semantic features.

**Criss-Cross Attention Module (CCAM).** As the core mechanism of CCNet, CCAM only calculates the pixels in the same row and column with strong relevance to the current pixel, which solves the problem of large consumption of global attention computation, and then performs multiple criss-cross attention computations to obtain the global dependency of each pixel by increasing the loop parameters, which solves the problem of sparse access to local attention information. To be specific, as shown in Fig. 3, after the feature extractor ResNet-101 network acquires the feature map F of the input image, it uses one convolution operation to perform a channel downscaling to obtain a new input feature $X$. Then, the feature map $X$ is updated to a new attentional feature map $X'$ using CCAM. The specific structure of CCAM is shown in Fig. 7. It first obtains feature maps $K$, $Q$ and $V$ from the input feature map $X$ through three 1 × 1 convolutional layers with unshared weights, then performs Affinity operation on $K$ and $Q$ and obtains spatial attention weight coefficients $A$ by softmax calculation, next performs Aggregation operation with the transformed feature map $V$, and finally sums with the input feature $X$ to obtain the attention feature $X'$ with aggregated contextual information. However, since CCAM only perceives the contextual information on the criss path, it is difficult to obtain dense and comprehensive information in a one-time calculation, and thus CCNet adds the cyclic parameter R to CCAM to form a second cycle, namely, cyclic criss-cross attention (RCCA). The parameters of RCCA are shared, and it takes the feature $X'$ generated for the first time using CCAM as input, and then generates a new feature map $X''$ through the information flow of the whole map, so that it can obtain rich contextual information of all pixels and ensure that the computational efficiency is not reduced.

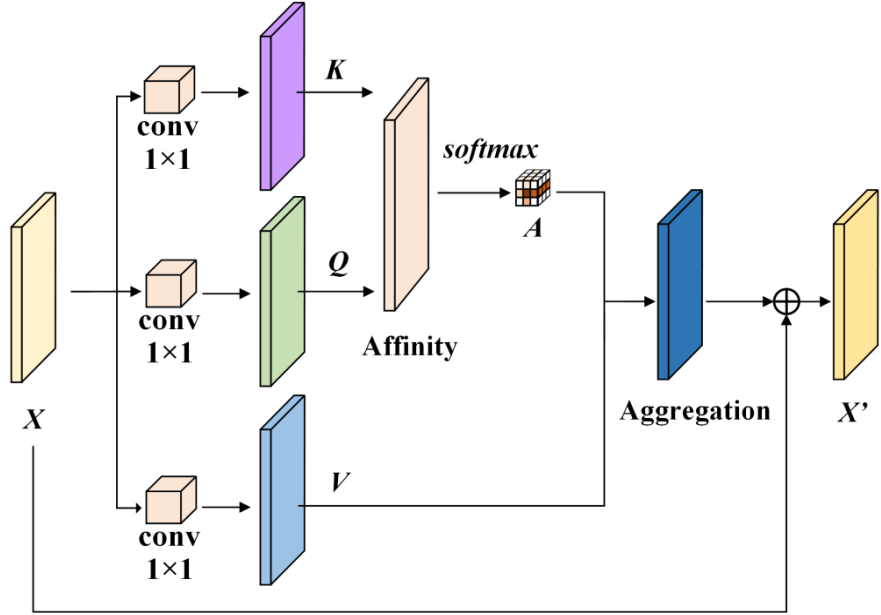

**Fig. 7.** Criss-Cross Attention Module.

**Semantic segmentation.** After obtaining the dense contextual features $X''$, they are connected to the original feature map $F$ and feature fusion is performed after batch normalization and activation operations in one convolutional layer. Then, the fused feature maps are mapped into each pixel to obtain the final crop prediction segmentation results.

**Change discriminator.** To acquire the change feature maps of crops in bi-temporal images, we consider BCD as a binary classification task. First, the features extracted from the T1 and T2 temporal images are differenced and the absolute values are taken to obtain the feature maps of the same size. Then, we use FCN to conduct the final binary segmentation to acquire the change map.

**Post-processing and constraint operations.** To obtain the initial semantic change map, we assign the T1 and T2 temporal crop segmentation results to each change pixel based on the binary change map for obtaining the semantic change prediction result of "T1 temporal crop to T2 temporal crop". In the no-change region, we also reserve the important no-change crop pixels as an additional semantic category. To obtain the final parcel-scale SCD results, we spatially constrain the initial crop SCD results with the fused bi-temporal farm parcel results according to the category area maximum rule within the parcel, so that each parcel has a unique semantic change category.

**Loss function.** In this study, we consider the SCD task as two sub-tasks of semantic segmentation and BCD, so we need to design a total loss function to weight the loss functions of the two sub-tasks in the image, and the formula is as follows.

$$Loss = Loss_{T1} + Loss_{T2} + 2 \times Loss_{BCD} \tag{1}$$

Where: the bi-temporal semantic segmentation part $Loss_{T1}$ and $Loss_{T2}$ both use the categorical cross-entropy loss function (Li et al. 2022a), and the BCD part $Loss_{BCD}$ uses the binary cross-entropy loss function (Li et al. 2021). The weight of $Loss_{BCD}$ is 2 in calculating the total $Loss$ because the no-change region of the bi-temporal images is the same.

## 2.4. Experimental details

### 2.4.1. Model implementation and training

**Software and hardware environment configuration.** This study conduct experiments on a computer configured with a 3.6 GHz Inter Core i7-9700K CPU, NVIDIA GeForce GTX 1080 Ti graphic card, 32G of RAM, Windows Server 2019 operating system, and PyTorch as the DL implementation framework.

**Dataset sample allocation.** In this study, we apply the CSCD dataset to farm plot extraction and crop SCD experiments, so we require dividing the samples into parcel samples containing only edge information and crop samples with multiple types of semantic changes, where the crop samples contain not only crop types of T1 and T2 temporal images, but also samples of binary changes and semantic changes of crops. We divide the dataset experimental area into two regions A1 and A2, each region contains corresponding training, validation and testing sample sets, and the sample allocation ratio is 6:1:3 in order.

**Model training parameters setting.** The DL model used has a training epoch of 20, a batch size set to 4, a learning rate of 0.001, a sigmoid function for the activation function of BDCN, a ReLU function for the activation function of the modified CCNet, and an optimization algorithm using stochastic gradient descent (SGD) with momentum set to 0.9 and weight decay set to 0.0001.

### 2.4.2. Evaluation metrics

In this study, the evaluation of the model includes two aspects: one is to examine the effect of different SCD methods on the accuracy of crop SCD results; and the other is the evaluation of module applicability as a way to determine the effect of different modules in the AGSPNet framework on the accuracy of crop SCD results from high-resolution UAV images. We classify the results of model detection into four cases: true positive (*TP*), false negative (*FN*), false positive (*FP*), and true negative (*TN*). *TP* is the number of pixels correctly detected as positive samples, *TN* is the number of pixels correctly detected as negative samples, *FP* is the number of pixels that mistakenly detect negative samples as positive samples, and *FN* is the number of pixels that miss detect positive samples as negative samples. We calculate the confusion matrix of SCD results with the true value of each pixel and use six accuracy evaluation metrics, precision (*Pre*), recall (*Rec*), F1-score (*F1*), overall accuracy (*OA*), kappa coefficient (*KC*) and mean intersection over union (*mIoU*), to evaluate the prediction performance of different benchmarks for SCD of crops from UAV high-resolution images. These evaluation metrics are able to check the model at various levels (e.g., accuracy, completeness, and consistency of results), and their values are proportional to the performance of the model. The specific formulas for the different accuracy evaluation metrics are as follows.

$$Pre = \frac{TP}{TP + FP} \tag{2}$$

$$Rec = \frac{TP}{TP + FN} \tag{3}$$

$$F1 = 2 \times \frac{Pre \times Rec}{Pre + Rec} \tag{4}$$

$$OA = \frac{TP + TN}{TP + FN + FP + TN} \tag{5}$$

$$R = \frac{(TP + FN) \times (TP + FP) + (TN + FN) \times (TN + FP)}{(TP + TN + FP + FN)^2} \tag{6}$$

$$KC = \frac{OA - R}{1 - R} \tag{7}$$

$$mIoU = \frac{TP}{FP + FN + TP} \tag{8}$$

Where: *R* is the process calculation variable of *KC*.

## 3. Experimental results

*3.1. Quantitative comparison of different semantic change detection methods*

To validate the performance of the proposed AGSPNet framework for crop SCD, we embed U-Net, PSPNet, DeepLabV3+, Mask R-CNN and HRNet in the SCD module of the framework separately, and conduct comparative experiments in test areas A1 and A2 of the CSCD dataset, which have scattered, varied and fragmented farmland distribution, surrounded by mountainous vegetation in the surrounding geography, and interspersed with lakes, roads and rural residences, which pose great challenges for crop SCD. The detailed and overall quantitative results of SCD of crops from bi-temporal images in A1 and A2 test areas by different DL methods are shown in Tables 2 and 3, respectively.

According to Table 2, it can be seen that on detailed crop SCD, due to the agro-geographic scene division module and parcel extraction module in the AGSPNet framework, different DL methods show stable and high performance results, and the SCD module (AGSPNet) used in this study achieves the highest F1 value for each crop semantic change detected. In particular, the agricultural geographic scene division module, by removing complex background features, enables the model to maintain the detection results of no changed background areas at a high accuracy level, with the highest F1 value reaching 0.989. Meanwhile, the parcel extraction module, by fusing pixel-level results into complete farm parcels conforming to the visual morphology, enables different SCD methods to achieve high accuracy detection results for the four crop semantic change categories as well, with AGSPNet having the highest average F1 values in A1 and A2 test areas, which are 0.033 and 0.037 higher than the sub-advanced method, respectively. In both test areas, the semantic change categories with the highest detection accuracy are concentrated in early-season rice to middle-season rice and early-season rice to late-season rice, which may be attributed to the fact that the change samples in these two categories are richer so that the models are adequately trained. The categories with the lowest detection accuracy are mainly vegetable and nursery with no change region, which is not only related to their low sample size, but also to their difficult to detect parcel boundaries and complex spatial distribution characteristics. Nevertheless, the proposed AGSPNet can still maintain a high detection accuracy in the above cases, mainly because of the powerful spatial context extraction capability of the criss-attention module.

**Table 2.** Quantitative comparison of different semantic change detection methods on each crop.

| Method | Change Type | A1 | | | A2 | | |
|---|---|---|---|---|---|---|---|
| | | *Pre* | *Rec* | *F1* | *Pre* | *Rec* | *F1* |
| U-Net | no change | 97.7% | 97.6% | 0.976 | 99.0% | 98.2% | 0.986 |
| | vegetable to vegetable | 80.1% | 81.5% | 0.808 | 79.7% | 79.2% | 0.794 |
| | nursery to nursery | 85.6% | 65.1% | 0.740 | 71.7% | 87.9% | 0.790 |
| | early-season rice to middle-season rice | 85.3% | 93.6% | 0.893 | 89.2% | 97.7% | 0.933 |
| | early-season rice to late-season rice | 94.3% | 93.6% | 0.939 | 93.9% | 96.6% | 0.952 |
| | rapeseed to middle-season rice | 92.6% | 85.7% | 0.890 | 95.4% | 80.6% | 0.874 |
| | rapeseed to late-season rice | 82.7% | 86.7% | 0.847 | 94.6% | 84.7% | 0.894 |
| PSPNet | no change | 97.2% | 97.8% | 0.975 | 98.9% | 98.3% | 0.986 |
| | vegetable to vegetable | 82.1% | 76.9% | 0.794 | 83.3% | 76.8% | 0.799 |
| | nursery to nursery | 87.8% | 45.1% | 0.596 | 64.8% | 84.2% | 0.732 |
| | early-season rice to middle-season rice | 85.9% | 91.3% | 0.885 | 91.2% | 96.7% | 0.939 |
| | early-season rice to late-season rice | 93.9% | 92.9% | 0.934 | 93.9% | 94.9% | 0.944 |
| | rapeseed to middle-season rice | 91.0% | 86.0% | 0.884 | 93.0% | 84.9% | 0.888 |
| | rapeseed to late-season rice | 79.5% | 86.0% | 0.826 | 87.9% | 85.5% | 0.867 |
| DeeplabV3+ | no change | 96.9% | 97.7% | 0.973 | 98.8% | 98.5% | 0.986 |
| | vegetable to vegetable | 84.0% | 64.6% | 0.730 | 80.6% | 82.1% | 0.813 |
| | nursery to nursery | 88.3% | 61.7% | 0.726 | 77.2% | 88.6% | 0.825 |
| | early-season rice to middle-season rice | 83.1% | 81.0% | 0.820 | 93.2% | 96.5% | 0.948 |
| | early-season rice to late-season rice | 88.0% | 94.0% | 0.909 | 94.7% | 95.0% | 0.948 |
| | rapeseed to middle-season rice | 93.7% | 71.6% | 0.812 | 93.6% | 89.6% | 0.916 |
| | rapeseed to late-season rice | 69.7% | 80.7% | 0.748 | 89.9% | 87.3% | 0.886 |
| Mask R-CNN | no change | 97.3% | 97.6% | 0.974 | 98.9% | 98.2% | 0.985 |
| | vegetable to vegetable | 82.4% | 83.8% | 0.831 | 80.1% | 86.3% | 0.831 |
| | nursery to nursery | 86.1% | 77.0% | 0.813 | 63.5% | 85.5% | 0.729 |
| | early-season rice to middle-season rice | 92.7% | 92.9% | 0.928 | 93.9% | 94.9% | 0.944 |
| | early-season rice to late-season rice | 95.7% | 89.1% | 0.923 | 92.7% | 96.1% | 0.944 |
| | rapeseed to middle-season rice | 93.3% | 96.3% | 0.948 | 91.8% | 89.9% | 0.908 |
| | rapeseed to late-season rice | 74.6% | 90.4% | 0.817 | 90.2% | 81.7% | 0.857 |
| HRNet | no change | 97.9% | 97.7% | 0.978 | 99.1% | 98.3% | 0.987 |
| | vegetable to vegetable | 80.9% | 83.2% | 0.820 | 77.2% | 80.2% | 0.787 |
| | nursery to nursery | 83.4% | 72.4% | 0.775 | 76.6% | 88.9% | 0.823 |
| | early-season rice to middle-season rice | 86.4% | 90.5% | 0.884 | 92.0% | 97.2% | 0.945 |
| | early-season rice to late-season rice | 94.4% | 93.8% | 0.941 | 94.1% | 96.3% | 0.952 |
| | rapeseed to middle-season rice | 90.0% | 88.1% | 0.890 | 94.2% | 88.3% | 0.912 |
| | rapeseed to late-season rice | 84.0% | 86.4% | 0.852 | 92.8% | 86.0% | 0.893 |
|  | no change | 97.4% | 98.0% | 0.977 | 99.2% | 98.6% | 0.989 |
|  | vegetable to vegetable | 86.3% | 80.0% | 0.830 | 85.4% | 86.4% | 0.859 |
|  | nursery to nursery | 81.4% | 83.0% | 0.822 | 88.4% | 92.0% | 0.902 |
| AGSPNet | early-season rice to middle-season rice | 95.8% | 94.9% | 0.953 | 96.3% | 97.1% | 0.967 |
|  | early-season rice to late-season rice | 94.4% | 93.0% | 0.937 | 95.1% | 98.2% | 0.966 |
|  | rapeseed to middle-season rice | 91.7% | 91.1% | 0.914 | 95.6% | 94.3% | 0.949 |
|  | rapeseed to late-season rice | 94.1% | 95.0% | 0.945 | 96.7% | 96.4% | 0.965 |

From Table 3, it is clear that on the overall crop SCD, corresponding to the detailed quantitative results, the DL model can assign more loss optimization parameters to multiple crop semantic change categories due to the reduced area share of complex backgrounds, while the parcel-scale detection results compensate to a greater extent for the deficiencies of the semantic segmentation algorithm, resulting in higher evaluation metrics for all the different SCD methods. In particular, the SCD accuracy of AGSPNet is still at the highest value, with F1, KC, OA and mIoU in the two test areas improving 0.038, 0.021, 0.011 and 0.062, respectively, on average over the sub-advanced algorithm HRNet. Whereas, compared with other methods, the detection accuracy of all three classical SSNs, U-Net, PSPNet and DeeplabV3+, is relatively low, which may be related to their poor detail information extraction ability, making it difficult to obtain localization-accurate and fine-grained edge segmentation results.

**Table 3.** Quantitative comparison of different semantic change detection methods on overall crop.

| Method | A1 | | | | | | A2 | | | | | |
|---|---|---|---|---|---|---|---|---|---|---|---|---|
|  | *Pre* | *Rec* | *F1* | *KC* | *OA* | *mIOU* | *Pre* | *Rec* | *F1* | *KC* | *OA* | *mIOU* |
| U-Net | 88.3% | 86.2% | 0.872 | 0.909 | 0.948 | 0.778 | 89.1% | 89.3% | 0.892 | 0.934 | 0.965 | 0.807 |
| PSPNet | 88.2% | 82.3% | 0.851 | 0.900 | 0.944 | 0.743 | 87.6% | 88.7% | 0.881 | 0.931 | 0.964 | 0.794 |
| DeeplabV3+ | 86.2% | 78.7% | 0.823 | 0.869 | 0.926 | 0.700 | 89.7% | 91.1% | 0.904 | 0.940 | 0.968 | 0.829 |
| Mask R-CNN | 88.9% | 89.6% | 0.892 | 0.911 | 0.950 | 0.809 | 87.3% | 90.4% | 0.888 | 0.932 | 0.964 | 0.804 |
| HRNet | 88.1% | 87.5% | 0.878 | 0.911 | 0.949 | 0.787 | 89.4% | 90.7% | 0.900 | 0.939 | 0.968 | 0.824 |
| AGSPNet | 91.6% | 90.7% | 0.911 | 0.931 | 0.961 | 0.842 | 93.8% | 94.7% | 0.942 | 0.961 | 0.979 | 0.894 |

*3.2. Visual comparison of different semantic change detection methods*

Fig. 8 and Fig. 9 show the visual comparison results of the different SCD methods for multiple crop semantic change categories in the A1 and A2 test areas. It can be obviously noted that the visualization results of the crop semantic changes further validate the analysis of the quantitative comparisons very well. After removing the complex background features of non-agriculture and fusing the parcel-level output results, the crop semantic change results detected by each SCD method obtain a parcel boundary shape that approximates the reference data, which ensures that the detected crop semantic change regions have a highly complete edge outline. Notably, the proposed AGSPNet is overall better than the other methods for crop SCD visualization in different test areas, especially with the best accuracy for early-season rice to middle-season rice and early-season rice to late-season rice detection, while vegetable and nursery in the no-change area also retain better boundary detail information, which explains their higher *Pre* in Table 2. The results of HRNet, PSPNet and U-Net produce more false detections on early-season rice to late-season rice (e.g., (d), (e) and (h) in Fig. 8 and Fig. 9), while the results of Mask R-CNN have more missed detections on early-season rice to middle-season rice (e.g., (g) in Fig. 8 and Fig. 9). The visualization of the DeeplabV3+ result distribution is the worst, with a large number of false detections of rapeseed to late-season rice and missed detections of rapeseed to middle-season rice (e.g., Fig. 8 (f)).

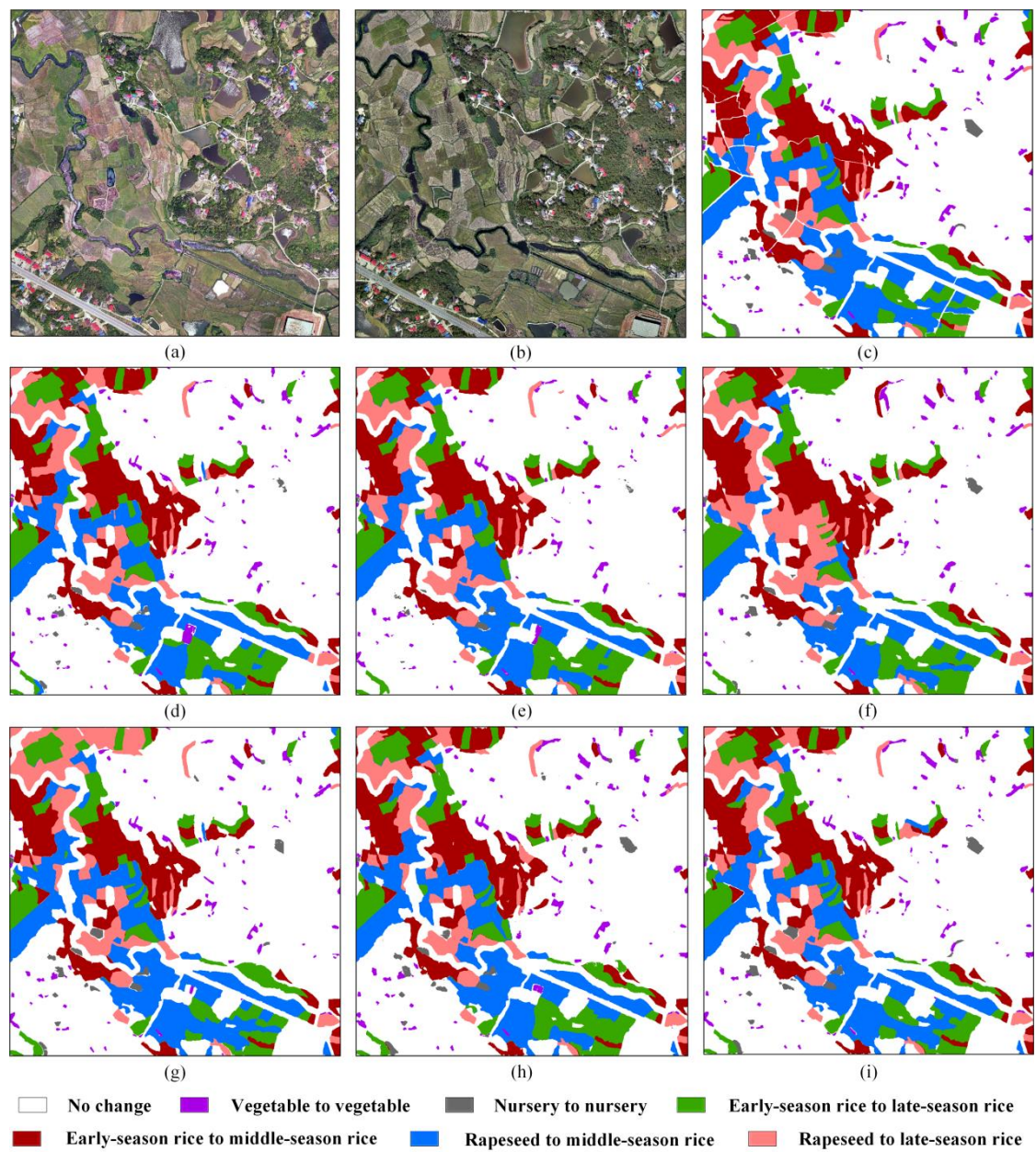

**Fig. 8.** Results of different semantic change detection methods in A1 test area. (a) T1 Image; (b) T2 Image; (c) ground truth; (d) U-Net; (e) PSPNet; (f) DeeplabV3+; (g) Mask R-CNN; (h) HRNet; (i) AGSPNet.

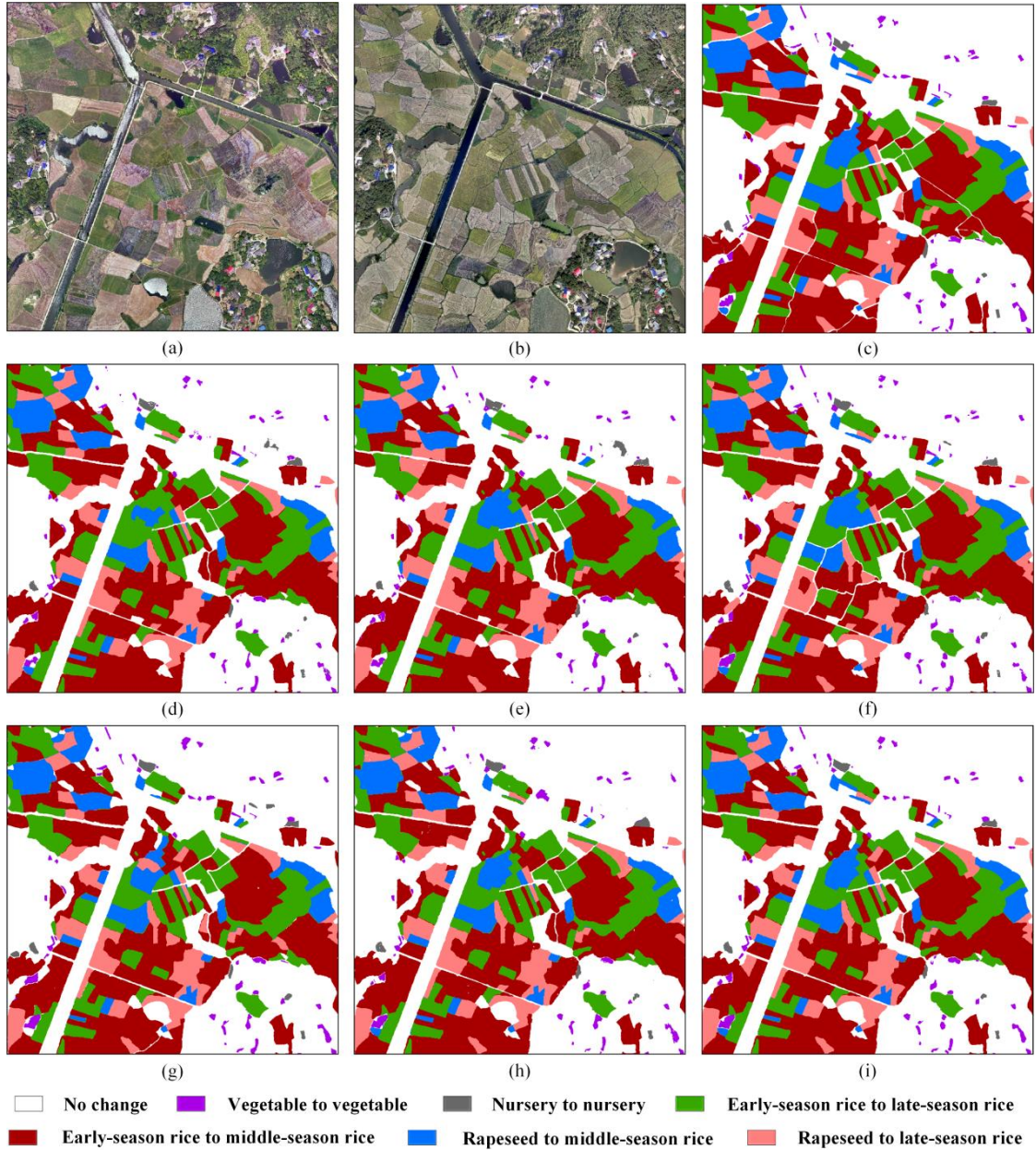

**Fig. 9.** Results of different semantic change detection methods in A2 test area. (a) T1 Image; (b) T2 Image; (c) ground truth; (d) U-Net; (e) PSPNet; (f) DeeplabV3+; (g) Mask R-CNN; (h) HRNet; (i) AGSPNet.

## 4. Discussion

### 4.1. Ablation experiments

#### 4.1.1. Quantitative comparison of different modules

To further validate the effects of different modules of AGSPNet on crop SCD results, we conduct ablation studies in test areas A1 and A2. Table 4 and Table 5 report the detailed and overall quantitative comparison results of crop SCD with different module combinations, respectively, where we denote the SCD module used in this study as the baseline model BASE, the agrogeographic scene delineation module as AGS, the plot edge extraction module as BDCN, and the AGSPNet framework as the combination of all modules (BASE + AGS + BDCN). From Table 4, it is observed that without the constraints of AGS and BDCN, the detection accuracy of the BASE model for each crop semantic change type in both A1 and A2 test areas decreases markedly, with the average *F1* values of background, vegetable and nursery in the no-semantic-change area decreasing to 0.97, 0.8 and 0.421, respectively, while the main four crop semantic change categories decrease the average *F1* value by 0.075 compared to that of the AGSPNet framework. When only the constraint of agricultural geographic scenes is added, the detection results of BASE + AGS in different categories of no-semantic-change regions have obvious accuracy improvement, and the *F1* value is improved by 0.13 on average over the BASE results, which indicates that the exclusion of complex non-agricultural background features is beneficial to improve the extraction effect of SCD model for no-change regions with relatively consistent geographic element conditions. As for the detection of semantic change regions of crops, the accuracy of BASE + AGS results has limited improvement, which may be owing to the pixel-level results of the semantic segmentation algorithm limiting its complete detection on crop boundaries. When only the constraints of parcel boundaries are added, the results of BASE + BDCN obtain a substantial improvement in the detection accuracy for both crop types with semantic changes, and the *F1* value increases by 0.035 on average over the results of BASE. This indicates that fusing the SCD results with the parcel-scale results of edge extraction can effectively improve the defects of the semantic segmentation algorithm and detect more complete and accurate crop parcels. However, this combination of modules still interferes with complex background features, resulting in a small improvement in the detection accuracy of no-semantic-change regions. When all modules are combined together, the model achieves higher accuracy gains than the above module combinations for the vast majority of change types, with *F1* values for regions with no-semantic-change averaging 0.154 higher than those of BASE + BDCN, and *F1* values for crop semantic change categories averaging 0.046 higher than those of BASE + AGS. This indicates that the AGSPNet framework is able to aggregate the advantages of AGS and BDCN modules, and the detected SCD results are less affected by the changeless complex background and "salt-and-pepper" noise.

**Table 4.** Quantitative comparison of different modules on each crop.

| Baseline | Change Type | A1 | | | A2 | | |
|---|---|---|---|---|---|---|---|
| | | *Pre* | *Rec* | *F1* | *Pre* | *Rec* | *F1* |
| BASE | no change | 96.7% | 97.3% | 0.970 | 98.7% | 95.3% | 0.970 |
| | vegetable to vegetable | 75.5% | 80.3% | 0.778 | 83.9% | 79.7% | 0.817 |
| | nursery to nursery | 56.4% | 59.3% | 0.578 | 15.8% | 80.5% | 0.264 |
| | early-season rice to middle-season rice | 93.1% | 92.0% | 0.925 | 94.2% | 93.7% | 0.939 |
| | early-season rice to late-season rice | 87.2% | 86.7% | 0.869 | 92.8% | 91.4% | 0.921 |
| | rapeseed to middle-season rice | 76.8% | 82.0% | 0.793 | 83.6% | 86.8% | 0.852 |
| | rapeseed to late-season rice | 90.2% | 81.9% | 0.858 | 83.0% | 87.8% | 0.853 |
| BASE + AGS | no change | 97.4% | 98.3% | 0.978 | 99.0% | 98.5% | 0.987 |
| | vegetable to vegetable | 87.2% | 81.8% | 0.844 | 86.0% | 78.8% | 0.822 |
| | nursery to nursery | 76.7% | 71.1% | 0.738 | 67.7% | 88.2% | 0.766 |
| | early-season rice to middle-season rice | 94.8% | 93.1% | 0.939 | 94.8% | 96.0% | 0.954 |
| | early-season rice to late-season rice | 89.0% | 86.5% | 0.877 | 92.5% | 94.1% | 0.933 |
| | rapeseed to middle-season rice | 82.2% | 85.6% | 0.839 | 91.2% | 88.3% | 0.897 |
| | rapeseed to late-season rice | 89.4% | 88.3% | 0.888 | 88.6% | 88.6% | 0.886 |

| | | | | | | | | |
|---|---|---|---|---|---|---|---|---|
| BASE + BDCN | no change | 96.9% | 97.2% | 0.970 | 98.9% | 96.3% | 0.976 |
| | vegetable to vegetable | 81.4% | 80.0% | 0.807 | 80.7% | 83.0% | 0.818 |
| | nursery to nursery | 65.5% | 61.4% | 0.634 | 22.5% | 81.0% | 0.352 |
| | early-season rice to middle-season rice | 92.4% | 94.0% | 0.932 | 94.8% | 96.7% | 0.957 |
| | early-season rice to late-season rice | 87.4% | 90.4% | 0.889 | 93.8% | 95.9% | 0.948 |
| | rapeseed to middle-season rice | 84.2% | 79.1% | 0.816 | 93.1% | 88.9% | 0.910 |
| | rapeseed to late-season rice | 91.1% | 85.9% | 0.884 | 92.7% | 91.8% | 0.922 |
| BASE + AGS + BDCN | no change | 97.4% | 98.0% | 0.977 | 99.2% | 98.6% | 0.989 |
| | vegetable to vegetable | 86.3% | 80.0% | 0.830 | 85.4% | 86.4% | 0.859 |
| | nursery to nursery | 81.4% | 83.0% | 0.822 | 88.4% | 92.0% | 0.902 |
| | early-season rice to middle-season rice | 95.8% | 94.9% | 0.953 | 96.3% | 97.1% | 0.967 |
| | early-season rice to late-season rice | 94.4% | 93.0% | 0.937 | 95.1% | 98.2% | 0.966 |
| | rapeseed to middle-season rice | 91.7% | 91.1% | 0.914 | 95.6% | 94.3% | 0.949 |
| | rapeseed to late-season rice | 94.1% | 95.0% | 0.945 | 96.7% | 96.4% | 0.965 |

From Table 5, it can be found that AGPSNet, which incorporates all modules, achieves the best accuracy for crop SCD results in both A1 and A2 test areas as a whole, with the largest improvement in the evaluation metric *mIoU*, which is 0.072 higher than the sub-advanced baseline on average, indicating that the structure we designed optimizes the detection results of the SCD model to the maximum extent possible. Moreover, the accuracy of BASE + AGS detection in the module combination is overall better than that of BASE + BDCN, where the average accuracy of six evaluation metrics is 0.025 higher, which indicates that the sample disproportionation problem caused by large unvarying areas may seriously inhibit the improvement of SCD model accuracy.

**Table 5.** Quantitative comparison of different modules accuracy on overall crop.

| Baseline | A1 | | | | | | A2 | | | | | |
|---|---|---|---|---|---|---|---|---|---|---|---|---|
| | *Pre* | *Rec* | *F1* | *KC* | *OA* | *mIoU* | *Pre* | *Rec* | *F1* | *KC* | *OA* | *mIoU* |
| BASE | 82.3% | 82.8% | 0.825 | 0.882 | 0.934 | 0.718 | 78.9% | 87.9% | 0.832 | 0.885 | 0.938 | 0.716 |
| BASE + AGS | 88.1% | 86.4% | 0.872 | 0.909 | 0.949 | 0.78 | 88.5% | 90.4% | 0.894 | 0.937 | 0.967 | 0.813 |
| BASE + BDCN | 85.6% | 84.0% | 0.848 | 0.894 | 0.941 | 0.748 | 82.4% | 90.5% | 0.863 | 0.919 | 0.957 | 0.767 |
| BASE + AGS + BDCN | 91.6% | 90.7% | 0.911 | 0.931 | 0.961 | 0.842 | 93.8% | 94.7% | 0.942 | 0.961 | 0.979 | 0.894 |

*4.1.2. Visual comparison of different modules*

Fig. 10 and Fig. 11 report the crop SCD visualization results for different module combinations in test areas A1 and A2, respectively. It can be evidently observed that the module combination of BASE + AGS + BDCN achieves the best visualization results for crop SCD in both test areas, detecting semantic change regions and no-change regions of crops accurately and comprehensively (e.g., the second row of Fig. 10 (g)), while being able to finely depict the gaps between crop parcels (e.g., the first and fourth rows of Fig. 11 (g)). The results of BASE + AGS improve the incorrect detection of complex mountain vegetation as nursery by BASE (as shown in the second row of Fig. 11), and the crop areas where semantic changes occur are also detected more accurately than by BASE. Although generating less "salt-and-pepper" noise than BASE, this combination of modules still struggles to ensure the internal integrity and edge accuracy of crop parcels. The results of BASE + BDCN can obtain comprehensive and complete crop parcels, but it also has more crop semantic change type detection mistakes affected by irrelevant background features, such as the large misdetection of early-season rice to late-season rice in the first row of Fig. 10 (f), the omission of Rapeseed to late-season rice in the fourth row, and the misdetection of nursery in the second row of Fig. 11 (f).

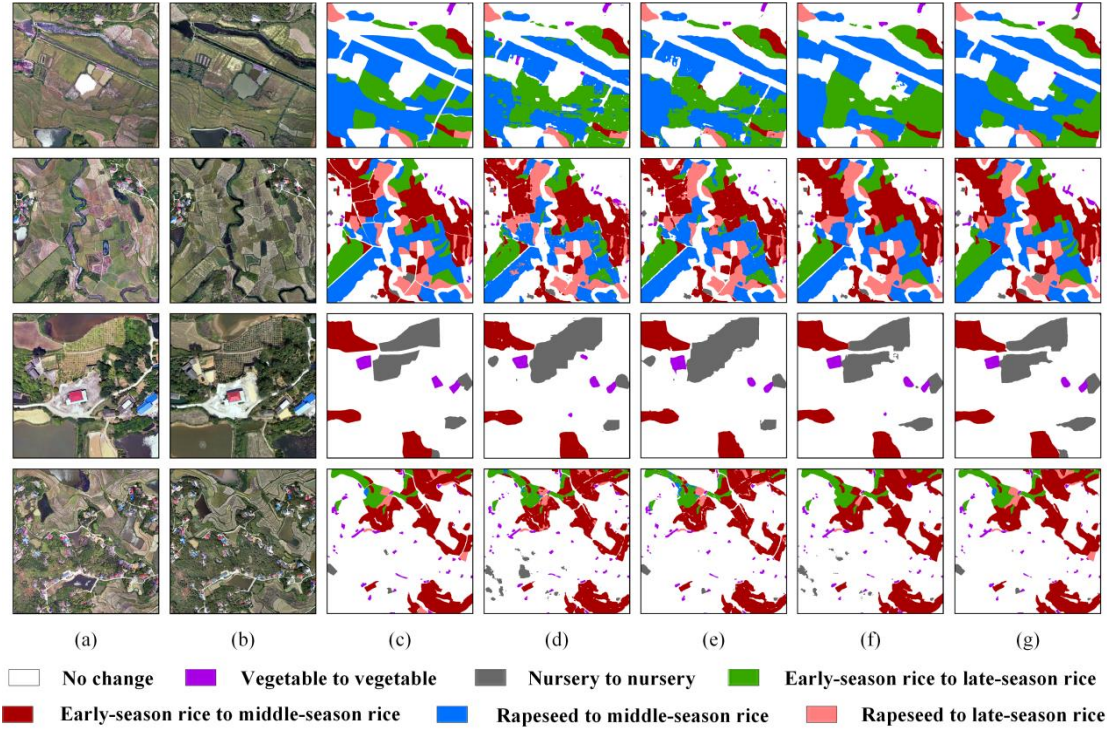

**Fig. 10.** Results of semantic change detection for different modules in A1 test area. (a) T1 Image; (b) T2 Image; (c) ground truth; (d) BASE; (e) BASE + AGS; (f) BASE + BDCN; (g) BASE + AGS + BDCN.

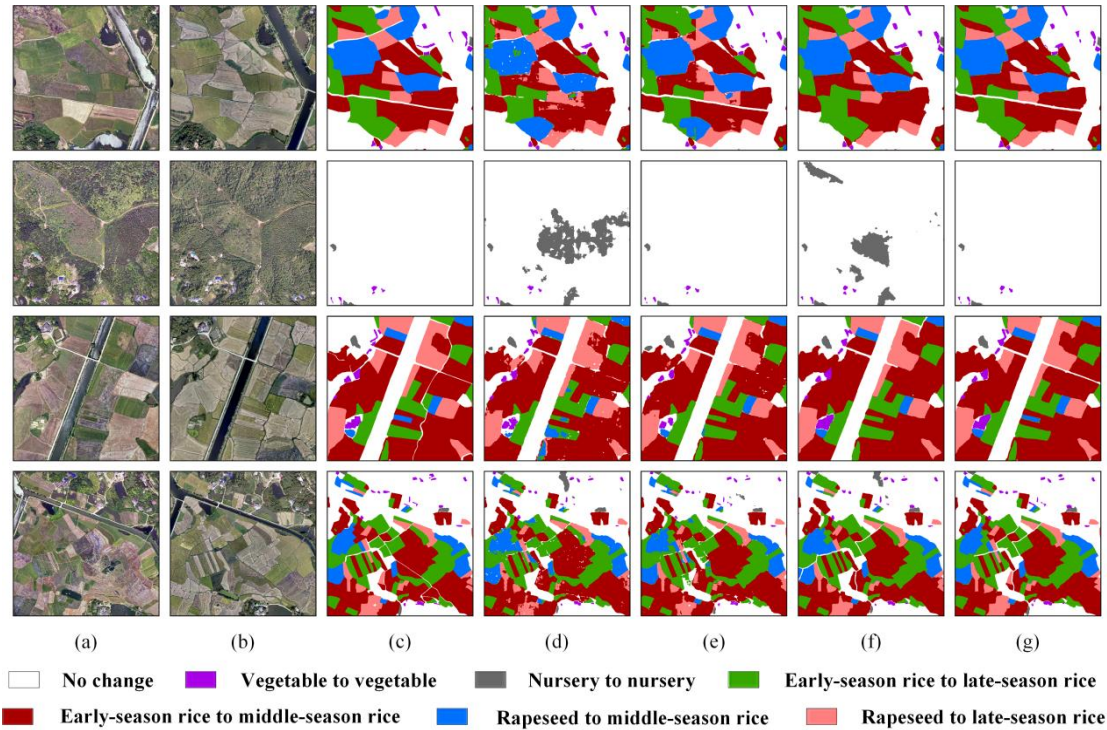

**Fig. 11.** Results of semantic change detection for different modules in A2 test area. (a) T1 Image; (b) T2 Image; (c) ground truth; (d) BASE; (e) BASE + AGS; (f) BASE + BDCN; (g) BASE + AGS +

BDCN.

*4.2. Uncertainty analysis*

Compared to existing work, our framework has several advantages. (1) It considers the problem of geospatial heterogeneity, reduces the interference of changeless background regions in crop SCD through the division of agricultural geographic scenes, and improves the detection of details of different semantic change types in crops. (2) It ensures the completeness of the farmland detection area by outputting parcel-scale crop semantic change results, and improves the problem of parcel fragmentation and severe "salt-and-pepper" noise caused by semantic segmentation algorithms. (3) It uses the pseudo-Siamese structure of CCNet and change feature discrimination module, which effectively reduces the spectral confusion detection and enhances the extraction capability of the model for important temporal features and spatial context information. In the experimental comparison of the proposed CSCD dataset, our framework demonstrates superior crop SCD performance over other advanced DL methods.

However, owing to the embeddable multi-task combination framework, the crop SCD results of AGSPNet may be affected by structural uncertainties within the different modules in the following aspects.

(1) The quality of public source geographic data affects the final delineation of agricultural geographic scene. If the data is of low quality, it may lead to the erroneous exclusion of areas with target objects in the process of geographic analysis, such as land cover products with low regional accuracy. Thus, the selection of multi-source geographic data products requires detailed validation and comparison to ensure that the data meet the quality demand for high accuracy.

(2) The parcel edge extraction results affect the accuracy of crop SCD results. If the extracted parcel edges are inaccurate, the incorrect semantic change categories may be output at the parcel scale, resulting in error accumulation. Therefore, the most suitable DL edge extraction model is required to be selected according to the actual application scenarios.

(3) Uncoordinated sample proportions affect the training effect of SCD models. In this study, due to the small sample area ratio of vegetable and nursery (Table 1), it is often difficult to accurately mine their features during network training, making the final SCD results much less accurate than other crop semantic change categories (Table 2). This problem can be improved in the future by investigating few-shot learning or data augmentation methods.

**5. Conclusion**

In this paper, we propose a crop SCD framework (AGSPNet) based on UAV bi-temporal high-resolution RSIs with agricultural geographic scene and parcel-scale constraints, and create a new open-source multi-class crop SCD dataset CSCD. The AGSPNet framework first gradually divides the agricultural geographic scenes with relatively consistent topographic features in UAV images based on the rule of agricultural territorial differentiation, so as to facilitate the depth feature extraction of complex crops by the DL model. Then, we introduce the edge detection network BDCN and design the parcel edge optimization model for extracting the complete parcels of crops in the agricultural geographic scenes. Finally, we fuse the pseudo-Siamese structure of CCNet with the change feature discrimination module, and output the obtained pixel-level SCD results with optimized parcel edges as spatial constraints to output parcel-scale crop semantic change results, thus alleviating the problems of inaccurate crop localization, incomplete targets and unclear edges of traditional SCD algorithms.

Meanwhile, the produced CSCD dataset focuses on yearly fine-grained agricultural monitoring, which largely complements the application scope and depth of the existing SCD dataset. We validate the effectiveness of AGSPNet on the CSCD dataset, and the results show that the AGSPNet crop semantic change results obtain the best performance in both quantitative metrics assessment and visual comparison compared to other SCD methods. The proposed framework can cope with a variety of practical and complex scenarios agricultural applications, detect more comprehensive and accurate crop fine-grained change areas, and provide technical reference for the development of smart agriculture and precision agriculture.

**CRediT authorship contribution statement**

**Shaochun Li:** conceptualization, methodology, software, validation, formal analysis, investigation, resources, writing—original draft preparation, writing—review and editing, supervision. **Yangjun Wang:** conceptualization, methodology, project administration. **Hengfan Cai:** methodology, resources. **Lina Deng:** data curation. **Yunhao Lin:** data curation.

**Declaration of competing interest**

The authors declare that they have no known competing financial interests or personal relationships that could have appeared to influence the work reported in this paper.

**Acknowledgment**

This research was funded by the National Natural Science Foundation of China (Nos. 41971423 and 31972951).

**Data availability**

The CSCD dataset is available at https://doi.org/10.6084/m9.figshare.22561537.v1.

**References**

Cai, Y., Guan, K., Peng, J., Wang, S., Seifert, C., Wardlow, B. & Li, Z. 2018. A high-performance and in-season classification system of field-level crop types using time-series Landsat data and a machine learning approach. *Remote Sens. Environ.*, **210**, 35-47. http://doi.org/10.1016/j.rse.2018.02.045.

Daudt, R.C., Le Saux, B., Boulch, A. & Gousseau, Y. 2019. Multitask learning for large-scale semantic change detection. *Comput. Vision Image Understanding*, **187**, 102783. http://doi.org/10.1016/j.cviu.2019.07.003.

Fan, X., Yan, C., Fan, J. & Wang, N. 2022. Improved U-Net Remote Sensing Classification Algorithm Fusing Attention and Multiscale Features. *Remote Sens.*, **14**, 3591. http://doi.org/10.3390/rs14153591.

Gerhards, M., Schlerf, M., Mallick, K. & Udelhoven, T. 2019. Challenges and future perspectives of multi-/Hyperspectral thermal infrared remote sensing for crop water-stress detection: A review. *Remote Sens.*, **11**, 1240. http://doi.org/10.3390/rs11101240.

He, J., Zhang, S., Yang, M., Shan, Y. & Huang, T. 2019. Bi-directional cascade network for perceptual edge detection. *Proc. IEEE/CVF Conf. Comput. Vision Pattern Recognit.*, 3828-3837.

Huang, Z., Wang, X., Huang, L., Huang, C., Wei, Y. & Liu, W. 2019. Ccnet: Criss-cross attention for semantic segmentation. *Proc. IEEE/CVF Int. Conf. Comput. Vision*, 603-612.

Hussain, M., Chen, D., Cheng, A., Wei, H. & Stanley, D. 2013. Change detection from remotely sensed images: From pixel-based to object-based approaches. *ISPRS J. Photogramm. Remote Sens.*, **80**, 91-106. http://doi.org/10.1016/j.isprsjprs.2013.03.006.

Kalinicheva, E., Ienco, D., Sublime, J. & Trocan, M. 2020. Unsupervised change detection analysis in satellite image time series using deep learning combined with graph-based approaches. *IEEE J. Sel. Top. Appl. Earth Obs. Remote Sens.*, **13**, 1450-1466. http://doi.org/10.1109/JSTARS.2020.2982631.

Li, H., Qiu, K., Chen, L., Mei, X., Hong, L. & Tao, C. 2020. SCAttNet: Semantic segmentation network with spatial and channel attention mechanism for high-resolution remote sensing images. *IEEE Geosci. Remote Sens. Lett.*, **18**, 905-909. http://doi.org/10.1109/LGRS.2020.2988294.

Li, P., He, X., Cheng, X., Qiao, M., Song, D., Chen, M., Zhou, T., Li, J*., et al.* 2022a. An improved categorical cross entropy for remote sensing image classification based on noisy labels. *Expert Syst. Appl.*, **205**, 117296. http://doi.org/10.1016/j.eswa.2022.117296.

Li, W., Yang, C., Peng, Y. & Du, J. 2022b. A pseudo-siamese deep convolutional neural network for spatiotemporal satellite image fusion. *IEEE J. Sel. Top. Appl. Earth Obs. Remote Sens.*, **15**, 1205-1220. http://doi.org/10.1109/JSTARS.2022.3143464.

Li, X., He, M., Li, H. & Shen, H. 2021. A combined loss-based multiscale fully convolutional network for high-resolution remote sensing image change detection. *IEEE Geosci. Remote Sens. Lett.*, **19**, 1-5. http://doi.org/10.1109/LGRS.2021.3098774.

Liang, S., Hua, Z. & Li, J. 2022. Transformer-based multi-scale feature fusion network for remote sensing change detection. *J. Appl. Remote Sens.*, **16**, 046509. http://doi.org/10.1117/1.JRS.16.046509.

Liu, M., Chai, Z., Deng, H. & Liu, R. 2022. A CNN-transformer network with multiscale context aggregation for fine-grained cropland change detection. *IEEE J. Sel. Top. Appl. Earth Obs. Remote Sens.*, **15**, 4297-4306. http://doi.org/10.1109/JSTARS.2022.3177235.

Liu, M., Shi, Q., Marinoni, A., He, D., Liu, X. & Zhang, L. 2021. Super-resolution-based change detection network with stacked attention module for images with different resolutions. *IEEE Trans. Geosci. Remote Sens.*, **60**, 1-18. http://doi.org/10.1109/TGRS.2021.3091758.

Liu, W., Wang, J., Luo, J., Wu, Z., Chen, J., Zhou, Y., Sun, Y., Shen, Z*., et al.* 2020. Farmland parcel mapping in mountain areas using time-series SAR data and VHR optical images. *Remote Sens.*, **12**, 3733. http://doi.org/10.3390/rs12223733.

Mesquita, D.B., dos Santos, R.F., Macharet, D.G., Campos, M.F. & Nascimento, E.R. 2019. Fully convolutional siamese autoencoder for change detection in UAV aerial images. *IEEE Geosci. Remote Sens. Lett.*, **17**, 1455-1459. http://doi.org/10.1109/LGRS.2019.2945906.

Myint, S.W., Gober, P., Brazel, A., Grossman-Clarke, S. & Weng, Q. 2011. Per-pixel vs. object-based classification of urban land cover extraction using high spatial resolution imagery. *Remote Sens. Environ.*, **115**, 1145-1161. http://doi.org/10.1016/j.rse.2010.12.017.

Patra, R.K., Patil, S.N., Falkowski-Gilski, P., Łubniewski, Z. & Poongodan, R. 2022. Feature Weighted Attention—Bidirectional Long Short Term Memory Model for Change Detection in Remote Sensing Images. *Remote Sens.*, **14**, 5402. http://doi.org/10.3390/rs14215402.

Peng, D., Zhang, Y. & Guan, H. 2019. End-to-end change detection for high resolution satellite images using improved UNet++. *Remote Sens.*, **11**, 1382. http://doi.org/10.3390/rs11111382.

Peng, D., Bruzzone, L., Zhang, Y., Guan, H. & He, P. 2021. SCDNET: A novel convolutional network for semantic change detection in high resolution optical remote sensing imagery. *Int. J. Appl. Earth Obs. Geoinf.*, **103**, 102465. http://doi.org/10.1016/j.jag.2021.102465.

Peng, D., Bruzzone, L., Zhang, Y., Guan, H., Ding, H. & Huang, X. 2020. SemiCDNet: A semisupervised convolutional neural network for change detection in high resolution remote-sensing images. *IEEE Trans. Geosci. Remote Sens.*, **59**, 5891-5906. http://doi.org/10.1109/TGRS.2020.3011913.

Shafique, A., Cao, G., Khan, Z., Asad, M. & Aslam, M. 2022. Deep learning-based change detection in remote sensing images: A review. *Remote Sens.*, **14**, 871. http://doi.org/10.3390/rs14040871.

Shi, Q., Liu, M., Li, S., Liu, X., Wang, F. & Zhang, L. 2021. A deeply supervised attention metric-based network and an open aerial image dataset for remote sensing change detection. *IEEE Trans. Geosci. Remote Sens.*, **60**, 1-16. http://doi.org/10.1109/TGRS.2021.3085870.

Song, A. & Choi, J. 2020. Fully convolutional networks with multiscale 3D filters and transfer learning for change detection in high spatial resolution satellite images. *Remote Sens.*, **12**, 799, http://doi.org/10.3390/rs12050799.

Sun, W., Sheng, W., Zhou, R., Zhu, Y., Chen, A., Zhao, S. & Zhang, Q. 2022. Deep edge enhancement-based semantic segmentation network for farmland segmentation with satellite imagery. *Comput. Electron. Agric.*, **202**, 107273. http://doi.org/10.1016/j.compag.2022.107273.

Sun, Y., Luo, J., Xia, L., Wu, T., Gao, L., Dong, W., Hu, X. & Hai, Y. 2020. Geo-parcel-based crop classification in very-high-resolution images via hierarchical perception. *Int. J. Remote Sens.*, **41**, 1603-1624. http://doi.org/10.1080/01431161.2019.1673916.

Tobler, W.R. 1970. A computer movie simulating urban growth in the Detroit region. *Econ. Geogr.*, **46**, 234-240.

Wang, H., Chen, X., Zhang, T., Xu, Z. & Li, J. 2022. CCTNet: Coupled CNN and transformer network for crop segmentation of remote sensing images. *Remote Sens.*, **14**, 1956. http://doi.org/10.3390/rs14091956.

Wang, L., Wang, L., Wang, Q. & Atkinson, P.M. 2021a. SSA-SiamNet: Spectral–spatial-wise attention-based Siamese network for hyperspectral image change detection. *IEEE Trans. Geosci. Remote Sens.*, **60**, 1-18. http://doi.org/10.1109/TGRS.2021.3095899.

Wang, Y., Gao, L., Hong, D., Sha, J., Liu, L., Zhang, B., Rong, X. & Zhang, Y. 2021b. Mask DeepLab: End-to-end image segmentation for change detection in high-resolution remote sensing images. *Int. J. Appl. Earth Obs. Geoinf.*, **104**, 102582. http://doi.org/10.1016/j.jag.2021.102582.

Woodcock, C.E., Loveland, T.R., Herold, M. & Bauer, M.E. 2020. Transitioning from change detection to monitoring with remote sensing: A paradigm shift. *Remote Sens. Environ.*, **238**, 111558. http://doi.org/10.1016/j.rse.2019.111558.

Wu, B., Zhang, M., Zeng, H., Tian, F., Potgieter, A.B., Qin, X., Yan, N., Chang, S*., et al.* 2023. Challenges and opportunities in remote sensing-based crop monitoring: a review. *Nat. Sci. Rev.*, **10**, nwac290. http://doi.org/10.1093/nsr/nwac290.

Wu, C., Du, B., Cui, X. & Zhang, L. 2017. A post-classification change detection method based on iterative slow feature analysis and Bayesian soft fusion. *Remote Sens. Environ.*, **199**, 241-255. http://doi.org/10.1016/j.rse.2017.07.009.

Xu, L., Ming, D., Zhou, W., Bao, H., Chen, Y. & Ling, X. 2019. Farmland extraction from high spatial resolution remote sensing images based on stratified scale pre-estimation. *Remote Sens.*, **11**, 108. http://doi.org/10.3390/rs11020108.

Yang, K., Xia, G.-S., Liu, Z., Du, B., Yang, W., Pelillo, M. & Zhang, L. 2021. Asymmetric siamese networks for semantic change detection in aerial images. *IEEE Trans. Geosci. Remote Sens.*, **60**, 1-18. http://doi.org/10.1109/TGRS.2021.3113912.

Yu, W., Zhou, W., Qian, Y. & Yan, J. 2016. A new approach for land cover classification and change analysis: Integrating backdating and an object-based method. *Remote Sens. Environ.*, **177**, 37-47. http://doi.org/10.1016/j.rse.2016.02.030.

Yuan, P., Zhao, Q., Zhao, X., Wang, X., Long, X. & Zheng, Y. 2022. A transformer-based Siamese

network and an open optical dataset for semantic change detection of remote sensing images. *Int. J. Digital Earth*, **15**, 1506-1525. http://doi.org/10.1080/17538947.2022.2111470.

Zerrouki, N., Harrou, F., Sun, Y. & Hocini, L. 2019. A machine learning-based approach for land cover change detection using remote sensing and radiometric measurements. *IEEE Sensors J.*, **19**, 5843-5850. http://doi.org/10.1109/JSEN.2019.2904137.

Zhong, Y., Zhu, Q. & Zhang, L. 2015. Scene classification based on the multifeature fusion probabilistic topic model for high spatial resolution remote sensing imagery. *IEEE Trans. Geosci. Remote Sens.*, **53**, 6207-6222. http://doi.org/10.1109/TGRS.2015.2435801.

Zhu, Q., Guo, X., Li, Z. & Li, D. 2022a. A review of multi-class change detection for satellite remote sensing imagery. *Geo-Spat. Inf. Sci.*, 1-15. http://doi.org/10.1080/10095020.2022.2128902.

Zhu, Q., Guo, X., Deng, W., Shi, S., Guan, Q., Zhong, Y., Zhang, L. & Li, D. 2022b. Land-use/land-cover change detection based on a Siamese global learning framework for high spatial resolution remote sensing imagery. *ISPRS J. Photogramm. Remote Sens.*, **184**, 63-78. http://doi.org/10.1016/j.isprsjprs.2021.12.005.